# Intelligent Semantic Web Search Engines: A Brief Survey

G.Madhu[1] and Dr.A.Govardhan[2] Dr.T.V.Rajinikanth[3]

[1] G.Madhu, Sr.Asst. Professor, Dept of Information Technology, VNR VJIET, Hyderabad-90,A.P. INDIA.
E-mail: madhu_g@vnrvjiet.in
[2] Dr. A.Govardhan, Principal, J.N.T.U College of Engineering, Jagityal. Karimnagar Dist-505 452. A.P.INDIA
E-mail: govardhan_cse@yahoo.co.in
[3] Dr. T.V. RajiniKanth,, Professor & HOD, Dept of Information Tech, GRIET,Hyderabad-500072., A.P INDIA
E-mail: rajinitv_03@yahoo.co.in

*ABSTRACT*

*The World Wide Web (WWW) allows the people to share the information (data) from the large database repositories globally. The amount of information grows billions of databases. We need to search the information will specialize tools known generically search engine. There are many of search engines available today, retrieving meaningful information is difficult. However to overcome this problem in search engines to retrieve meaningful information intelligently, semantic web technologies are playing a major role. In this paper we present survey on the search engine generations and the role of search engines in intelligent web and semantic search technologies.*

## 1. INTRODUCTION

The Semantic Web is an extension of the current Web [1] that allows the meaning of information to be precisely described in terms of well-defined vocabularies that are understood by people and computers. On the Semantic Web information is described using a new W3C standard called the Resource Description Framework (RDF). Semantic Web Search is a search engine for the Semantic Web. Current Web sites can be used by both people and computers to precisely locate and gather information published on the Semantic Web. Ontology [2] is one of the most important concepts used in the semantic web infrastructure, and RDF(S) (Resource Description Framework/Schema) and OWL (Web Ontology Languages) are two W3C recommended data representation models which are used to represent ontologies. The Semantic Web will support more efficient discovery, automation, integration and reuse of data and provide support for interoperability problem which can not be resolved with current web technologies. Currently research on semantic web search engines are in the beginning stage, as the traditional search engines such as *Google, Yahoo, and Bing (MSN)* and so forth still dominate the present markets of search engines.

Most of the search engines search for keywords to answer the queries from users. The search engines usually search web pages for the required information. However they filter the pages from searching unnecessary pages by using advanced algorithms. These search engines can answer topic wise queries efficiently and effectively by developing state-of art algorithms. However they are vulnerable in answering intelligent queries from the user due to the dependence of their results on information available in web pages. The main focus of these search engines is solving these queries with close to accurate results in small time using much researched algorithms. However, it shows that such search engines are vulnerable in answering intelligent queries using this approach. They either show inaccurate results with this approach or show accurate but (could be) unreliable results. With the keywords based searches they usually provide results from blogs (if available) or other discussion boards. The user cannot have a satisfaction with these results due to lack of trusts on blogs etc. To overcome this problem in search engines to retrieve relevant and meaningful information intelligently, semantic web technology deals with a great role [3]. Intelligent semantic technology gives the nearer to desired results by search engines to the user.

In this paper, we will make a preliminary survey over the existing literature regarding intelligent semantic search engines and semantic web search. By classifying the literature into few main categories, we review their characteristics respectively. In addition, the issues within the reviewed intelligent semantic search methods and engines are analyzed and concluded based on perspectives.

## 2. BACKGROUND

Information retrieval by searching information on the web is not a fresh idea but has different challenges when it is compared to general information retrieval. Different search engines return different search results due to the variation in indexing and search process. Google, Yahoo, and Bing have been out there which handles the queries after processing the keywords. They only search information given on the web page, recently, some research group's start delivering results from their semantics based search engines, and however most of them are in their initial stages. Till none of the search engines come to close indexing the entire web content, much less the entire Internet.

Current web is the biggest global database that lacks the existence of a semantic structure and hence it makes difficult for the machine to understand the information provided by the user. When the information was distributed in web, we have two kinds of research problems in search engine i.e.

- How can a search engine map a query to documents where information is available but does not retrieve in intelligent and meaning full information?
- The query results produced by search engines are distributed across different documents that may be connected with hyperlink. How search engine can recognize efficiently such a distributed results?

Semantic web [4] [5], can solve the first problem in web with semantic annotations to produce intelligent and meaningful information by using query interface mechanism and ontology's. Other one can be solved by the graph-based query models [6]. The Semantic web would require solving extraordinarily difficult problems in the areas of knowledge representation, natural language understanding. The following figure depicts the semantic web frame work it also referred as the semantic web layercake by W3C.

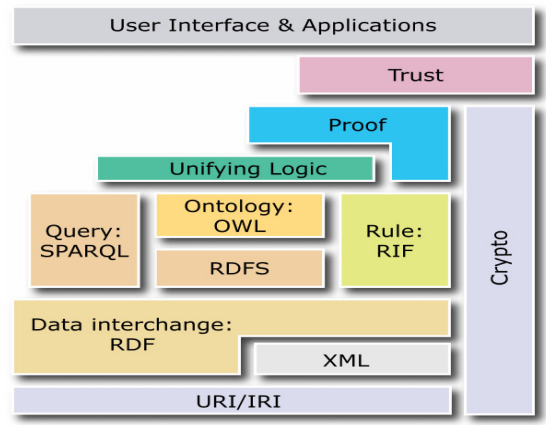

*Fig.1. Semantic Web Frame Work*

**2.1 Current Web & Limitations**

Present World Wide Web is the longest global database that lacks the existence of a semantic structure and hence it becomes difficult for the machine to understand the information provided by the user in the form of search strings. As for results, the search engines return the ambiguous or partially ambiguous result data set; Semantic web is being to be developed to overcome the following problems for current web.

- The web content lacks a proper structure regarding the representation of information.
- Ambiguity of information resulting from poor interconnection of information.
- Automatic information transfer is lacking.
- Usability to deal with enormous number of users and content ensuring trust at all levels.
- Incapability of machines to understand the provided information due to lack of a universal format.

Hakia [7] is a general purpose semantic search engine that search structured text like Wikipedia. Hakia calls itself a "meaning-based (semantic) search engine" [8]. They're trying to provide search results based on meaning match, rather than by the popularity of search terms. The presented news, Blogs, Credible, and galleries are processed by hakia's proprietary core semantic technology called QDEXing [7]. It can process any kind of digital artifact by its Semantic Rank technology using third party API feeds [9].

## 3. INTELLIGENT SEMANTIC WEB

*3.1 Intelligent Search Engines*

Currently, a couple of Intelligent search engines are designed and implemented for different working environments, and the mechanisms that realize these search engine are distinct.
*Fu-Ming Hung* and *Jenn-Hwa Yang* present an *intelligent search engine* with semantic technologies. This research has combine *description logic* inference system and digital library ontology to complete *intelligent search engine* [10]. According to search engine mechanism, presenting demands and a formula evaluating present related technology of that can solve and promote the efficiency of search engine, and formulating the demands of wisdom search engine. If uses *Description Logic Inference System* to integrate the digital library ontology to proceed

36

with the inference of user requirement, and combines the content search mechanism and knowledge inference to accomplish the study of *intelligent search engine.*

*Inamdar and Shinde* [11] discussed agent based *intelligent search engine* system for web mining. Most of the web search engines make use of the text only on a web page. Agents are used to perform some action or activity on behalf of a user of a computer system. Each user is assisted by his/her own personal agent to search the web. The major goal of each personal agent is to propose to its user and to other agent's links to web pages that are considered relevant for their search. Personal agents can use different internal and external sources of information. The personal agents are software agents running on the server [12].

*Patrick Lambrix and Nahid Shahmehri and Niclas Wahllof* [13] presents a search engine is described as one that tackles the problem of enhancing the precision and recall for retrieval of documents. The main techniques that they apply here are the use of subsumption information and the use of default information. The use of subsumption information allows for the retrieval of documents that include information about the desired topic as well as information about more specific topics. The use of default information allows for retrieving of documents that include typical content information about a topic. The strict and default information are represented in an extension of description logics that can deal with defaults. There have been tested the system on small-scale databases with promising results.

*Satya Sai Prakash* et al, present architecture and design specifications for new generation search engines highlighting the need for intelligence in search engines and give a knowledge framework to capture intuition. Simulation methodology to study the search engine behavior and performance is described. Simulation studies are conducted using fuzzy satisfaction function and heuristic search criterion after modeling client behavior and web dynamics [14].

*Dan Meng, Xu Huang* discussed an interactive intelligent search engine model based on user information preference [15]. This model can be an effective and useful way to realize the individuation information search for different user information preference. This model frame work, used some artificial intelligent methods and technologies to improve the quality and effectiveness of information retrieval.

*Xiajiong Shen  Yan Xu  Junyang Yu  Ke Zhang* forward an intelligent search engine where Information Retrieval model is found on formal context of FCA (formal concept analysis) and incorporates with a browsing mechanism for such a system based on the concept lattice. Test data validates its feasibility, and implement of the FCA-search engine indicates that the concept lattice of FCA is a useful way of supporting the flexible management of documents according to conceptual relation [16].

## 4. TYPES OF SEMANTIC SEARCH ENGINES

Semantic is the process of communicating enough meaning to result in an action. A sequence of symbols can be used to communicate meaning, and this communication can then affect behavior. Semantics has been driving the next generation of the Web as the Semantic Web, where the focus is on the role of semantics for automated approaches to exploiting Web resources. 'Semantic' also indicates that the meaning of data on the web can be discovered not just by people, but also by computers. Then the Semantic Web was created to extend the web and make data easy to reuse everywhere.

Semantic web is being developed to overcome the following main limitations of the current Web [17]:

- The web content lacks a proper structure regarding the representation of information.
- Ambiguity of information resulting from poor interconnection of information.
- Automatic information transfer is lacking.
- Unable to deal with enormous number of users and content ensuring trust at all levels.
- Incapability of machines to understand the provided information due to lack of a universal format.

### 4.1 Semantic search engines

Currently many of semantic search engines are developed and implemented in different working environments, and these mechanisms can be put into use to realize present search engines.

*Alcides Calsavara and Glauco Schmidt* proposes and defines a novel kind of service for the semantic search engine. A semantic search engine stores semantic information about Web resources and is able to solve complex queries, considering as well the context where the Web resource is targeted, and how a semantic search engine may be employed in order to permit clients obtain information about commercial products and services, as well as about sellers and service providers which can be hierarchically organized [18]. Semantic search engines may seriously contribute to the development of electronic business applications since it is based on strong theory and widely accepted standards.

*Sara Cohen Jonathan Mamou et al* presented a semantic search engine for XML (XSEarch) [19].It has a simple query language, suitable for a naïve user. It returns semantically related document fragments that satisfy the user's query. Query answers are ranked using extended information-retrieval techniques and are generated in an order similar to the ranking. Advanced indexing techniques were developed to facilitate efficient implementation of XSEarch. The performance of the different techniques as well as the recall and the precision were measured experimentally. These experiments indicate that XSEarch is efficient, scalable and ranks quality results highly.

*Bhagwat and Polyzotis* propose a Semantic-based file system search engine- Eureka, which uses an inference model to build the links between files and a File Rank metric to rank the files according to their semantic importance [20]. Eureka has two main parts: a) crawler which extracts file from file system and generates two kinds of indices: keywords' indices that record the keywords from crawled files, and rank index that records the File Rank metrics of the files; b) when search terms are entered, the query engine will match the search terms with keywords' indices, and determine the matched file sets and their ranking order by an information retrieval-based metrics and File Rank metrics.

*Wang et al*. project a semantic search methodology to retrieve information from normal tables, which has three main steps: identifying semantic relationships between table cells; converting tables into data in the form of database; retrieving objective data by query languages [21]. The research objective defined by the authors is how to use a given table and a given domain knowledge to convert a table into a database table with semantics. The authors' approach is to denote the layout by layout syntax grammar and match these denotation with given templates which can be used to analyze the semantics of table cells. Then semantic preserving transformation is used to transform tables to database format.

*Kandogan et al*. develop a semantic search engine-Avatar, which combines the traditional text search engine with use of ontology annotations [22]. Avatar has two main functions: a) extraction and representation – by means of UIMA framework, which is a workflow consisting of a chain of annotators extracted from documents and stored in the annotation store; b) interpretation – a process of automatically transforming a keyword search to several precise searches. Avatar consists of two main parts: semantic optimizer and user interaction engine. When a query is entered into the former, it will output a list of ranked interpretations for the query; then the top-ranked interpretations are passed to the latter, which will display the interpretations and the retrieved documents from the interpretations.

*4.2 Ontology search engines*

*Maedche et al*. designed an integrated approach for ontology searching, reuse and update [23]. In its architecture, an ontology registry is designed to store the metadata about ontologies and ontology server stores the ontologies. The ontologies in distributed ontology servers can be created, replicated and evolved. Ontology metadata in ontology registry can be queried and registered when a new ontology is created. Ontology search in ontology registry is executed under two conditions -query-by-example is to restrict search fields and search terms, and query-by-term is to restrict the hyponyms of terms for search.

*Georges Gardarin et al.* discussed a SEWISE [24] is an ontology-based Web information system to support Web information description and retrieval. According to domain ontology, SEWISE can map text information from various Web sources into one uniform XML structure and make hidden semantic in text accessible to program. The textual information of interest is automatically extracted by Web Wrappers from various Web sources and then text mining techniques such as categorization and summarization are used to process retrieved text information.

## 5. SOME COMMON ISSUES

We have discussed a preliminary survey of the existing and dynamic area in *intelligent semantic search engines* and methods. Although we have not claimed this survey is comprehensive, some common issues in the current semantic search engines and methods are concluded as follows:

a) **Low precision and high recall**
   Some Intelligent semantic search engines cannot show their significant performance in improving precision and lowering recall. *In Ding's* semantic flash search engine, the resource of the search engine is based on the top-50 returned results from Google that is not a semantic search engine, which could be low precision and high recall [25].

b) **Identity intention of the user**
   User intention identification plays an important role in the intelligent semantic search engine. For example, in *chiung-Hon leon lee* introduced a method for analyzing the request terms to fit user intention, so that the service provided will be more suitable for the user [26].

International journal of Web & Semantic Technology (IJWesT) Vol.2, No.1, January 2011**c) Individual user patterns can be extrapolated to global users.**

In early search engine that offered disambiguation to search terms. A user could enter in a search term that was ambiguous (e.g., Java) and the search engine would return a list of alternatives (coffee, programming language, island in the South Seas).

**d) Inaccurate queries.**

We have user typically domain specific knowledge. And users don't include all potential Synonyms and variations in the query, actually user have a problem but aren't sure how to phrase.

## 6 CONCLUSIONS

In this paper, we make a brief survey of the existing literature regarding intelligent semantic search technologies. We review their characteristics respectively. In addition, the issues within the reviewed intelligent semantic search methods and engines are concluded based on four perspectives differentiations between designers and users' perceptions, static knowledge structure, low precision and high recall and lack of experimental tests.
In the future, our work will focus on the deeper and broader research in the field of intelligent semantic search, with the purpose of concluding the current situation of the field and promote the further development of intelligent semantic search engine technologies.

International journal of Web & Semantic Technology (IJWesT) Vol.2, No.1, January 2011

## Authors

**[1]G.Madhu** completed his Master degree in Mathematics from J.N.T.University, Hyderabad in 2000 and his M.Tech degree in computer science & engineering from J.N.T.University, Hyderabad, INDIA, in 2008. Now pursuing PhD in Computer Science and Engineering from J.N.T.University, Hyderabad. He is presently working as Sr. Assistant Professor in Information Technology Department at VNR VJIET Hyderabad. His current research interest includes ANN, data mining, rough sets, and semantic web. He is a professional member of Indian Society for Rough Sets, and ISTE.

**[2]Dr.A.Govardhan** did his BE in Computer Science and Engineering from Osmania University College of Engineering, Hyderabad,M.Tech from Jawaharlal Nehru University, Delhi and Ph.D from Jawaharlal Nehru Technological University, Hyderabad. He is presently working as Principal, JNTU Jagtial, Karimnagar, A.P, INDIA. He has 63 research publications at International/National Journals and Conferences. He is also a reviewer of research papers of various conferences. His areas of interest include Databases, Data Warehousing & Mining, Information Retrieval, Computer Networks, Image Processing and Object Oriented Technologies.

**[3]Dr. T.V.K.Rajinikanth** received M.Tech from Osmania University, Hyderabad in 2001, PhD from Osmania University, and Hyderabad in 2007. He is currently working as HOD, Dept of IT, at GRIET, Hyderabad, A.P.INDIA; He has several national & International publications and conferences. His research interest includes   Data warehouse & mining, Semantic Web, Spatial Data mining, ANN.